\pdfoutput=1

\documentclass[11pt]{article}

\usepackage[final]{acl}

\usepackage{times}
\usepackage{latexsym}
\usepackage{booktabs}

\usepackage[T1]{fontenc}

\usepackage[utf8]{inputenc}

\usepackage{microtype}

\usepackage{inconsolata}

\usepackage{graphicx}

\usepackage{url}
\usepackage{tikz}
\usepackage{tabularx}

\usepackage{adjustbox}
\usepackage{xcolor}
\usepackage{hyperref}
 \definecolor{darkblue}{rgb}{0, 0, 0.5}
  \hypersetup{colorlinks=true, citecolor=darkblue, linkcolor=darkblue, urlcolor=darkblue}

\usepackage{xstring}

\usepackage{color}

\title{Evidence-backed Fact Checking using RAG and Few-Shot In-Context Learning with LLMs}

\author{
 \textbf{Ronit Singhal\textsuperscript{1}},
 \textbf{Pransh Patwa\textsuperscript{2}},
 \textbf{Parth Patwa\textsuperscript{3}},
\\
 \textbf{Aman Chadha\textsuperscript{4,5*}},
 \textbf{Amitava Das\textsuperscript{6}},
\\
\\
 \textsuperscript{1}IIT Kharagpur, India,
 \textsuperscript{2}Aditya English Medium School, India,
 \textsuperscript{3}UCLA, USA, \\
 \textsuperscript{4}Stanford University, USA,
 \textsuperscript{5}Amazon GenAI, USA,
 \textsuperscript{6}University of South Carolina, USA
\\ 
\textsuperscript{1}ronit@kgpian.iitkgp.ac.in, \textsuperscript{2}pransh.patwa@aemspune.edu.in, \textsuperscript{3}parthpatwa@g.ucla.edu\\
\textsuperscript{4,5}hi@aman.ai, 
\textsuperscript{6}amitava@mailbox.sc.edu
\\
}

\begin{document}
\maketitle
\renewcommand{\thefootnote}{\fnsymbol{footnote}}
\footnotetext[1]{Work does not relate to position at Amazon.}
\renewcommand*{\thefootnote}{\arabic{footnote}}
\setcounter{footnote}{0}

\begin{abstract}
Given the widespread dissemination of misinformation on social media, implementing fact-checking mechanisms for online claims is essential. Manually verifying every claim is very challenging, underscoring the need for an automated fact-checking system. This paper presents our system designed to address this issue. We utilize the Averitec dataset \cite{schlichtkrull2023averitec} to assess the performance of our fact-checking system. In addition to veracity prediction, our system provides supporting evidence, which is extracted from the dataset. We develop a Retrieve and Generate (RAG) pipeline to extract relevant evidence sentences from a knowledge base, which are then inputted along with the claim into a large language model (LLM) for classification. We also evaluate the few-shot In-Context Learning (ICL) capabilities of multiple LLMs. Our system achieves an 'Averitec' score of 0.33, which is a 22\% absolute improvement over the baseline.  Our Code is publicly available on \href{https://github.com/ronit-singhal/evidence-backed-fact-checking-using-rag-and-few-shot-in-context-learning-with-llms}{https://github.com/ronit-singhal/evidence-backed-fact-checking-using-rag-and-few-shot-in-context-learning-with-llms}.

\end{abstract}

\section{Introduction}

The proliferation of fake news and misinformation on social media platforms has emerged as a significant contemporary issue \cite{fake_news_threat_to_society}. False online claims have, in some cases, incited riots \cite{riots} and even resulted in loss of life \cite{death}. This problem is particularly amplified during critical events such as elections \cite{Bovet2019} and pandemics \cite{iran_fake_news,bae2022challengesequitablevaccinedistribution,morales2021covid19testsgonerogue}. Given the vast volume of online content, manually fact-checking every claim is impractical. Therefore, the development of an automated fact verification system is imperative. Moreover, simply assigning a veracity label is inadequate; the prediction must be supported by evidence to ensure the system's transparency and to bolster public trust. Although recent solutions have been proposed \cite{patwa-fake,capuano2023content}, the problem remains far from resolved and requires further research efforts.

\begin{figure}[t]
    \centering
    \includegraphics[width=\linewidth]{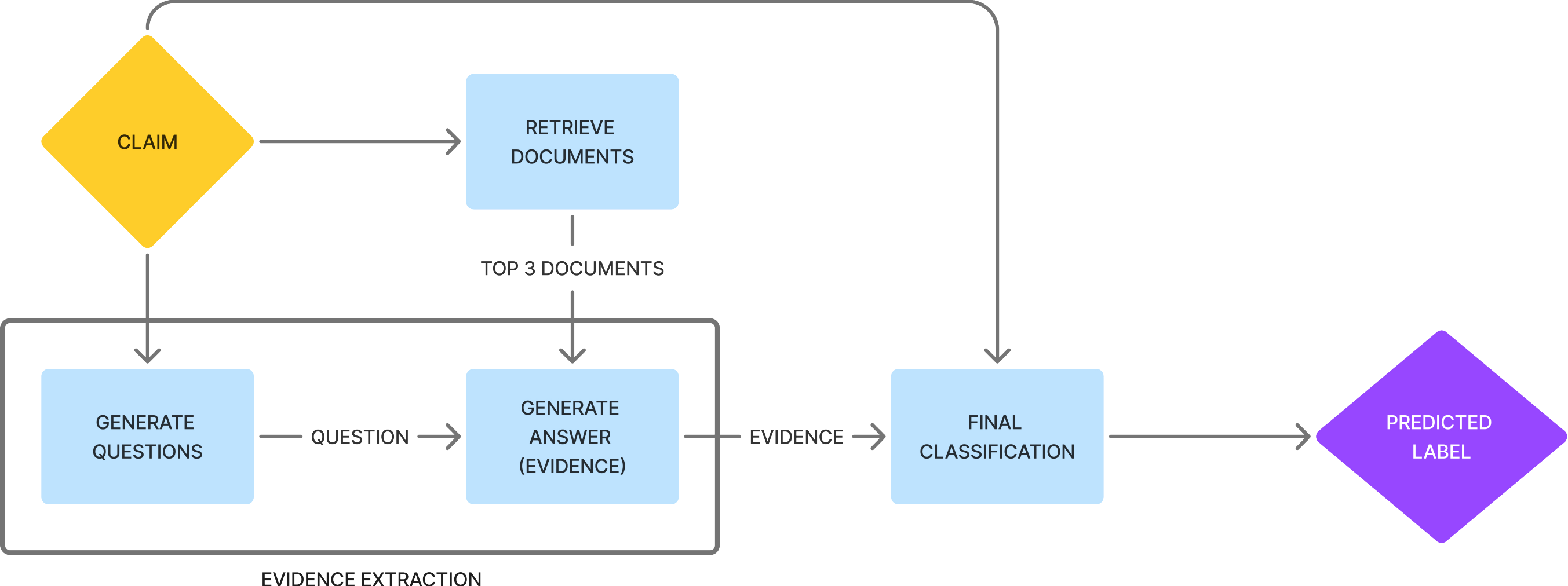}
    \caption{Overview diagram of our system. First, the claim is used to retrieve the top 3 relevant documents. Next, evidence is extracted from these documents using questions and answers generated by an LLM. Finally, the evidence is used for veracity prediction via few-shot ICL. }
    \label{fig:overview}
\end{figure}

In this paper, we present our system for automated fact verification. Our system classifies a given textual claim into one of four categories: Supported, Refuted, Conflicting Evidence/Cherrypicking, or Not Enough Evidence. Additionally, it provides supporting evidence for the classification. Our approach leverages recent advancements in Large Language Models (LLMs), specifically Retrieval-Augmented Generation (RAG) and In-Context Learning (ICL), to produce evidence-backed veracity predictions. Given a claim and a collection of documents, our system first employs a RAG pipeline to retrieve the three most relevant documents and extract evidence from them. Subsequently, we utilize ICL to determine the veracity of the claim based on the extracted evidence. Figure \ref{fig:overview} provides a high-level overview of our system. We evaluate our system on the Averitec dataset \cite{schlichtkrull2023averitec}, where it outperforms the official baseline by a large margin. Our key contributions are as follows:

\begin{itemize}
    \item We develop a system for automated fact verification that integrates RAG with ICL to provide evidence-based classifications.
    \item Our proposed system requires only a minimal number of training samples, thereby eliminating the need for a large manually annotated dataset.
    \item We conduct experiments with various recent LLMs and provide a comprehensive analysis of the results.
\end{itemize}

The remainder of this paper is structured as follows: Section \ref{sec:related} provides a literature review of related works, while Section \ref{sec:data} describes the dataset. In Section \ref{sec:method}, we outline our methodology, followed by a detailed account of the experimental setup in Section  \ref{sec:experiments}. Section  \ref{sec:results} presents and analyzes our results, and finally, we conclude in Section \ref{sec:conclusion}.

\section{Related Work}
\label{sec:related}

Recently, there has been increased research interest in fake news detection and fact checking. \citet{Glazkova_2021} proposed an ensemble of BERT \cite{devlin-etal-2019-bert} for Covid fake news \cite{10.1007/978-3-030-73696-5_3} detection. \citet{10.1145/3501401} employed deep learning techniques for fact checking in Arabic \cite{baly2018integrating}. \cite{SONG2021102712} tackled the problem of fake news detection using graph neural networks. The factify tasks \cite{mishra2022factify,suryavardan2023factify} aimed to detect multi-modal fake news. However, these systems only provide the veracity prediction without any evidence. 

On the FEVER dataset \cite{thorne-etal-2018-fever}, \citet{krishna-etal-2022-proofver}  designed a seq2seq model to generate natural logic-based inferences as proofs, resulting in SoTA performance on the dataset. \citet{schuster-etal-2021-get} released the VitaminC dataset and propose contrastive learning for fact verification. \citet{hu-etal-2022-dual} proposed a DRQA retriever \cite{chen2017readingwikipediaansweropendomain} based method for fact checking over unstructured information \cite{aly-etal-2021-fact}.  These systems provide evidence or explanation to back their predictions but they test the veracity of synthetic claims whereas we test real claims.

Some researchers have also used LLMs to tackle the problem. \citet{kim2024llmsproducefaithfulexplanations} leveraged multiple LLMs as agents to enhance the faithfulness of explanations of evidence for fact-checking. \citet{zhang2023llmbasedfactverificationnews} designed a hierarchical prompting method which directs LLMs to separate a claim into several smaller claims and then verify each of them progressively.

There have also been attempts to solve the problem using RAG. \citet{khaliq2024ragar} utilized multimodal LLMs with a reasoning method called chain of RAG to provide evidence based on text and image. \citet{deng2024cramcredibilityawareattentionmodification}
proposed a method to decrease misinformation in RAG pipelines by re-ranking the documents during retrieval based on a credibility score assigned to them. Similar to these systems, we also use RAG and LLMs in our solution. 
 
For more detailed surveys, please refer to \citet{thorne2018automated,kotonya2020explainable,guo2022survey}.

\section{Data}
\label{sec:data}
\begin{table}[t]
\centering
\resizebox{0.9\columnwidth}{!}{%
\begin{tabular}{lcc}
\toprule
                   \textbf{Class} & \textbf{Train} & \textbf{Dev} \\ 
                    \toprule
Supported              &    847   &  122   \\ 
Refuted              &   1743    &   305  \\ 
Conflicting evidence/Cherrypicking       &    196   &   38  \\ 
Not enough evidence &   282    &   35  \\ 
\midrule
\textbf{Total}      &  3068     & 500    \\ 
\bottomrule
\end{tabular}
}
\caption{Class-wise distribution of train and dev set of the dataset. The data is skewed towards the Refuted class.}
\label{tab:distribution}
\end{table}
\begin{figure}
    \centering
    \includegraphics[width=\linewidth]{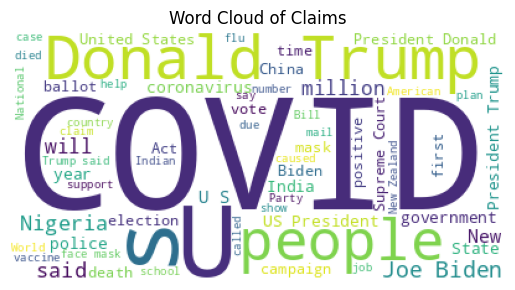}
    \caption{Word cloud of the claims. We can see that Politics and COVID-19 are common topics in the claims.}
    \label{fig:wordcloud}
\end{figure}

We utilize the Averitec dataset \cite{schlichtkrull2023averitec} for fact-checking purposes. This dataset comprises claims accompanied by a knowledge store (a collection of articles). Each claim is annotated with question-answer pairs that represent the evidence, a veracity label, and a justification for the label. The veracity label can be one of the following: Support (S), Refute (R), Conflicting Evidence/Cherrypicking (C), or Not Enough Evidence (N). A claim is labeled as C when it contains both supporting and refuting evidence. The data distribution, as shown in Table \ref{tab:distribution}, indicates a class imbalance favoring the R class, while the C and N classes have relatively few examples. The final testing is conducted on 2,215 instances \cite{averitec-overview}. For further details on the dataset, please refer to \citet{schlichtkrull2023averitec,averitec-overview}.

On average, each claim consists of 17 words. Figure \ref{fig:wordcloud} (word cloud of the claims) reveals that most claims are related to politics and COVID-19.

\section{Methodology}
\label{sec:method}

\begin{figure}[t]
    \centering
    \begin{tikzpicture}
    \node[draw, rounded corners] {
    \resizebox{0.75\linewidth}{!}{    
    \adjustbox{minipage=[r][27em][b]{0.4\textwidth},scale={0.7}}{
    \textbf{Convert the following claim to one neutral question. Do not miss out anything important from the claim. Question the claim, not the fact.}\\
    \textbf{\textcolor{blue}{Claim:}} Donald Trump has stated he will not contest for the next elections \\
    \textbf{\textcolor{blue}{Incorrect Question:}} "What did Donald Trump state for the next elections?"\\
     \textbf{\textcolor{blue}{Correct Question:}} "Did Donald Trump state that he will not contest for the next elections?" \\
    \textbf{\textcolor{blue}{Claim:}} [another claim]\\
    \textbf{\textcolor{blue}{Incorrect Question:}} [example of an incorrect question]\\
    \textbf{\textcolor{blue}{Correct Question:}} [expected correct question]\\
    \textbf{\textcolor{red}{Given claim:}} In a letter to Steve Jobs, Sean Connery refused to appear in an Apple commercial. \\
    \textbf{\textcolor{green}{Generated question:}} "Is it true that Sean Connery wrote a letter to Steve Jobs refusing to appear in an Apple commercial?" 
    }}
    };
    \end{tikzpicture}
    \caption{The prompt used for generating questions. Some manually created correct and incorrect examples are given to guide the LLM. }
    \label{fig:question_generate}
\end{figure}

\begin{figure}[t]
    \centering
    \begin{tikzpicture}
    \node[draw, rounded corners] {
    \resizebox{0.75\linewidth}{!}{    
    \adjustbox{minipage=[r][24.5em][b]{0.4\textwidth},scale={0.7}}{
    \textbf{Your task is to extract a portion of the provided text that directly answers the given question. The extracted information should be a conclusive answer, either affirmative or negative, and concise, without any irrelevant words. You do not need to provide any explanation. Only return the extracted sentence as instructed. You are strictly forbidden from generating any text of your own. }\\
    \textbf{\textcolor{red}{Question:}} Is it true that Sean Connery wrote a letter to Steve Jobs refusing to appear in an Apple commercial? \\
    \textbf{\textcolor{red}{Document text:}} [entire text of one of the retrieved documents] \\
     \textbf{\textcolor{green}{Generated answer:}} "No, it is not true that Sean Connery wrote a letter to Steve Jobs refusing to appear in an Apple commercial. The letter was a fabrication created for a satirical article on Scoopertino."
    }}
    };
    \end{tikzpicture}
    \caption{The prompt used for generating answers. This prompt is repeated for each of the top three documents.  }
    \label{fig:answer_generate}
\end{figure}

\begin{figure}[t]
    \centering
    \begin{tikzpicture}
    \node[draw, rounded corners] {
    \resizebox{0.9\linewidth}{!}{    
    \adjustbox{minipage=[r][33em][b]{0.5\textwidth},scale={0.7}}{
    \textbf{Classify the given claim based on provided statements into one of:  \\
    1. 'Supported' if there is sufficient evidence indicating that the claim is legitimate. 
    2. 'Refuted' if there is any evidence contradicting the claim. \\
    3. Not Enough Evidence' If you cannot find any conclusive factual evidence either supporting or refuting the claim.  \\ 
    4. 'Conflicting Evidence/Cherrypicking' if there is factual evidence both supporting and refuting the claim. 
    }\\
    \textbf{\textcolor{blue}{Claim:}} [claim ] \\
    \textbf{\textcolor{blue}{Statements:}} [statements related to claim] \\
    \textbf{\textcolor{blue}{Class:}} [ground truth class] \\
    \textbf{\textcolor{blue}{Claim:}} [claim] \\
    \textbf{\textcolor{blue}{Statements:}} [statements related to claim] \\
    \textbf{\textcolor{blue}{Class:}} [ground truth class] \\
     \textbf{\textcolor{red}{Given Claim:}} New Zealand's new Food Bill bans gardening. \\
    \textbf{\textcolor{red}{Given Statements:}} ["The Food Bill does not impose restrictions on personal horticultural activities, such as growing vegetables and fruits at home.", "Gardening is not banned in New Zealand.", "There are no laws against people having gardens, or sharing food that they've grown at home, said a spokesperson for New Zealand's Ministry for Primary Industries."]\\
    \textbf{\textcolor{green}{Generated class:}} Refuted
    }}
    };
    \end{tikzpicture}
    \caption{A prompt similar to the one used for generating the final prediction. The actual prompt has some more instructions which are omitted here in the interest of space. two annotated train examples are provided for the LLM to learn from. }
    \label{fig:class_generate}
\end{figure}

Given a claim and a knowledge store, our system is comprised of three key components: relevant document retrieval, evidence extraction from the documents, and veracity prediction based on the extracted evidence. The first two components form our Retrieval-Augmented Generation (RAG) pipeline.

\subsection{Document Retrieval Using Dense Embeddings}
In the document retrieval phase, it is essential to match claims with relevant documents from a knowledge store (in our case, the knowledge store consists of documents provided in the dataset, though it could be replaced with documents retrieved via a search engine). To facilitate this, all documents are first transformed into dense vector embeddings using an embedding model. Since our knowledge store is static, this transformation is a one-time process. The claim in question is then converted into embeddings using the same model.

Once the claim is embedded, we utilize FAISS (Facebook AI Similarity Search) \cite{douze2024faisslibrary} to conduct a nearest-neighbor search within the knowledge store. FAISS is an efficient library for similarity search and clustering of dense vectors. We configure FAISS to retrieve the top three documents most relevant to the claim. These documents are then used in the subsequent evidence extraction and veracity prediction steps.

\subsection{Evidence Extraction Using LLMs}
After identifying the top three relevant documents, the next step involves extracting evidence supported by these documents. This process consists of two steps:

\textbf{Question Generation}: The claim is transformed into a question challenging its validity using an LLM. We employ In-Context Learning, which enables the model to generate responses based on a few provided examples, aiding in the creation of nuanced and contextually appropriate questions. The prompt is designed to ensure that the generated question challenges the claim's veracity rather than simply seeking a factual answer. An example prompt is provided in Figure \ref{fig:question_generate}.

\textbf{Answer Generation}: After generating the question, we provide a single document to an LLM and pose the question. The LLM is prompted to deliver concise and definitive answers derived directly from the content of the document. This process is repeated for each of the three documents, resulting in three distinct answers for each claim. These answers collectively constitute our evidence. It is important to note that in our experiments, the LLM used for answer generation does not necessarily need to be the same as the one used for question generation. The prompt utilized in this step is similar to the one depicted in Figure \ref{fig:answer_generate}.

\subsection{Few-Shot ICL for Final Classification}

For the final veracity prediction, we use an LLM to classify a claim based on the three pieces of evidence extracted earlier. The LLM is prompted to choose one out of the four possible classes. The prompt is designed to guide the model through the classification process, ensuring that it correctly interprets the relationship between the claim and the evidence. An example prompt is given in Figure \ref{fig:class_generate}.

Our methodology aligns with recent advancements in retrieval-augmented generation (RAG) pipelines which alleviate hallucination and ICL methods, which have been shown to improve the accuracy of LLMs. The integration of these state-of-the-art methods is an attempt to ensure that the extracted evidence is both relevant and contextually appropriate  for validating the claims accurately. 

\section{Experiments}
\label{sec:experiments}
To convert documents into dense embeddings, we utilize the \texttt{dunzhang/stella\_en\_1.5B\_v5} model\footnote{\url{https://huggingface.co/dunzhang/stella_en_1.5B_v5}}. This model is chosen because, at the time of our experiments, it was ranked first on the Massive Text Embedding Benchmark (MTEB) leaderboard \cite{muennighoff2022mteb}, and holds the second position at the time of writing this paper.

For all LLMs used in our experiments, we employ their 4-bit quantized versions via Ollama\footnote{\url{https://github.com/ollama/ollama}}. This quantization enables us to load larger LLMs onto our GPUs.

For question generation, we use the Phi-3-medium model \cite{abdin2024phi3technicalreporthighly}. The temperature is set to 0, and greedy decoding is used to ensure that the answers are as factual as possible and to minimize hallucinations.

For answer generation and final classification, we experiment with multiple LLMs of varying sizes, including InternLM2.5 \cite{cai2024internlm2technicalreport}, Llama-3.1 \cite{dubey2024llama3herdmodels}, Phi-3-medium \cite{abdin2024phi3technicalreporthighly}, Qwen2 \cite{yang2024qwen2technicalreport}, and Mixtral \cite{jiang2024mixtralexperts}. These models are selected based on their performance on the Open LLM Leaderboard \cite{open-llm-leaderboard-v2} and their availability through Ollama.

We utilize an A40 GPU for Mixtral, while all other models are run on an A100 GPU. Our best-performing model, Mixtral, requires an average of 2 minutes for evidence extraction and final prediction. Our code is publicly available on \href{https://github.com/ronit-singhal/evidence-backed-fact-checking-using-rag-and-few-shot-in-context-learning-with-llms}{https://github.com/ronit-singhal/evidence-backed-fact-checking-using-rag-and-few-shot-in-context-learning-with-llms}.

\subsection{Evaluation Metrics}
\label{metrics}
The evaluation metrics used ensure that credit for a correct veracity prediction is given only when the correct evidence has been identified.

To evaluate how well the generated questions and answers align with the reference data, the pair-wise scoring function METEOR \cite{banerjee-lavie-2005-meteor} is used. The Hungarian Algorithm \cite{hungarian} is then applied to find the optimal matching between the generated sequences and the reference sequences. This evidence scoring method is referred to as Hungarian METEOR. The system is evaluated on the test set using the following metrics:

\begin{itemize}
    \item \textbf{Q only:} Hungarian METEOR score for the generated questions.
    \item \textbf{Q + A:} Hungarian METEOR score for the concatenation of the generated questions and answers.
    \item \textbf{Averitec Score:} Correct veracity predictions where the Q+A score is greater than or equal to 0.25. Any claim with a lower evidence score receives a score of 0.
\end{itemize}

\section{Results and Analysis}
\label{sec:results}

\begin{table}[t]
\centering
\resizebox{\columnwidth}{!}{%
\begin{tabular}{lcccc}
\toprule
\textbf{Model}        & \textbf{Size}  & \textbf{Q+A}   $\uparrow$         & \textbf{Averitec}  $\uparrow$      & \textbf{Acc}    $\uparrow$         \\ 
\toprule
InternLM2.5  & 7B    & 0.278          & 0.194          & 0.374          \\ 
Llama3.1     & 8B    & 0.259          & 0.224          & 0.538          \\ 
Phi-3-Medium & 14B   & 0.259          & 0.28           & 0.654          \\ 
Llama 3.1    & 70B   & 0.272          & 0.328          & \textbf{0.662} \\ 
Qwen2        & 72B   & 0.285          & 0.33           & 0.61           \\ 
Mixtral      & 8*22B & \textbf{0.292} & \textbf{0.356} & 0.636          \\ 
\bottomrule
\end{tabular}
}
\caption{Results of various models on the dev set. Performance improves as the model size increases. Acc refers to accuracy. Q+A and Averitec scores are described in Section \ref{metrics}.}
\label{tab:dev_models}
\end{table}

\begin{table}[t]
\begin{tabular}{lcccc}
\toprule
\textbf{System}   & Q  $\uparrow$  & Q+A $\uparrow$ & Averitec $\uparrow$ \\ \toprule
 Official Baseline  & 0.24 & 0.2 & 0.11     \\
 Mixtral (ours) & \textbf{0.35} & \textbf{0.27}  & \textbf{0.33}   \\\bottomrule 
\end{tabular}
\caption{Results on the test set. Our system which uses Mixtral for final prediction outperforms the official baseline in all metrics. For more details of the metrics, please refer to section \ref{metrics}. }
\label{tab:baseline}
\end{table}

Table \ref{tab:dev_models} provides a summary of the performance of various models on the development set. The Mixtral 8*22B model \cite{jiang2024mixtralexperts} achieves the highest Averitec score, while the Llama 3.1 model \cite{dubey2024llama3herdmodels} attains the highest accuracy. These findings indicate that model performance generally improves with increasing model size. Moreover, the relative rankings of these models on the development set differ from their positions on the Open LLM leaderboard \cite{open-llm-leaderboard-v2}, suggesting that superior performance on the Open LLM leaderboard does not necessarily correlate with better performance in the fact verification task.

Given that Mixtral achievs the highest Averitec score on the development set, we select it for evaluation on the test set. Table \ref{tab:baseline} provides a comparison of our system and the official baseline \cite{schlichtkrull2023averitec} on the test set. The baseline model utilizes Bloom \cite{bloom} for evidence generation, followed by re-ranking of the evidence using a finetuned BERT-large model and finally a finetuned BERT-large model veracity prediction.  Unlike the baseline, which uses finetuned models, we only use a few train examples via ICL. Despite that, our system outperforms the baseline across all three evaluation metrics. Notably, our Averitec score of 0.33 is a 22\% absolute improvement over the baseline.

\subsection{Class-wise Performance}

\begin{table}[t!]
\resizebox{\columnwidth}{!}{%
\begin{tabular}{lccccc}
\toprule
\textbf{Model}     & \textbf{S}              & \textbf{R}              & \textbf{N}              & \textbf{C}              & \textbf{Macro}         \\ 
\toprule
Mixtral   & 0.605          & 0.780          & 0.126          & 0.117          & \textbf{0.47} \\ 
Qwen2      & \textbf{0.620} & 0.754          & \textbf{0.157} & \textbf{0.153} & 0.42          \\ 
Llama 3.1 70b & 0.613          & \textbf{0.809} & 0.022          & 0              & 0.361         \\ 
\bottomrule
\end{tabular}
}
\caption{Class-wise F1 scores of our top three LLMs on the dev set. Classes are Supported (S), Refuted (R), Not enough evidence (N), and conflicting evidence/cherrypicking. Macro-averaged F1 score is also reported.  }
\label{tab:class-wise}
\end{table}

\begin{figure}[t!]
    \centering
    \includegraphics[width=\linewidth]{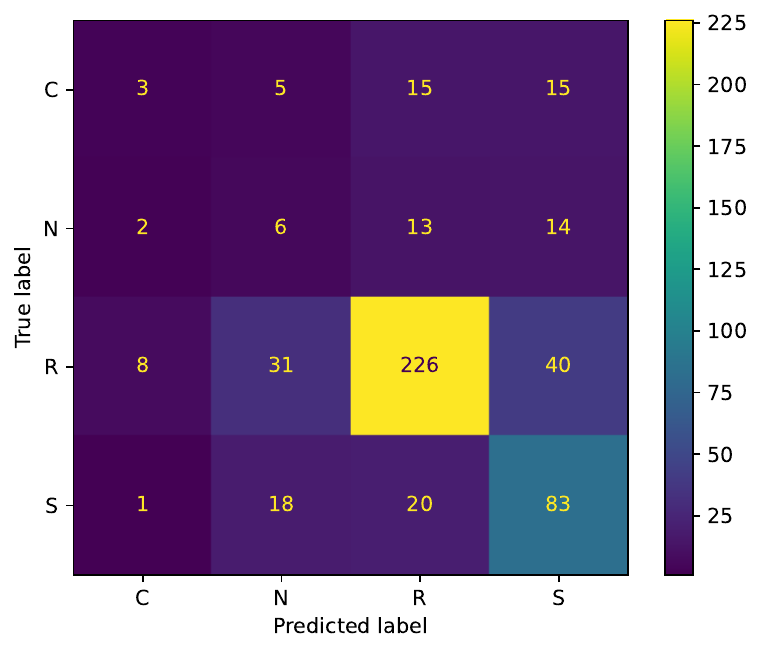}
    \caption{Confusion matrix of Mixtral on the development set, illustrating the model's performance across four classes (C, N, R, S). While class R is mostly accurately classified, classes C and N are often mis-predicted as R or N.}
    \label{fig:CM}
\end{figure}

Table \ref{tab:class-wise} presents the class-wise performance of our top three models on the development set. Across all models, the Refuted class emerges as the easiest to predict, while the "Not Enough Evidence" and "Conflicting Evidence/Cherrypicking" classes present greater challenges. Notably, no single model excels across all classes. Although Mixtral achieves the highest macro F1 score, it is not the top-performing model for any individual class. Qwen2 surpasses the other models in performance across all classes except Refuted. This suggests that exploring ensemble techniques could be a valuable direction for future research.

Figure \ref{fig:CM} illustrates the confusion matrix of Mixtral 8*22B on the development set. It reveals that both the N and C classes are equally likely to be misclassified as the R and S classes. Additionally, there is significant confusion between the S and R classes, highlighting the inherent difficulty of fact verification.

\section{Conclusion and Future Work}
\label{sec:conclusion}
In this paper, we introduced our system for evidence-supported automated fact verification. Our system - based on RAG and ICL - requires only a minimal number of training examples to extract relevant evidence and make veracity predictions. We observed that all LLMs demonstrate sub-optimal performance on the "Conflicting Evidence/Cherrypicking" and "Not Enough Evidence" categories, which emphasizes the inherent challenges of these categories. Additionally, no single LLM consistently outperforms others across all categories. Our system achieved an Averitec score of 0.33, highlighting the complexity of the problem and indicating a substantial potential for future improvement.

Future research could involve fine-tuning the LLM using parameter-efficient fine-tuning (PEFT) techniques \cite{liu2022fewshotparameterefficientfinetuningbetter,patwa-etal-2024-enhancing} and improving performance through the use of ensemble techniques \cite{mohammed2022effective}. Extending the system to include multi-modal fact verification \cite{factify1overview,suryavardan2023findingsfactify2multimodal} also represents an interesting direction for further investigation.

\section{Limitation}
As we are using few-shot ICL, our system cannot make use of large annotated datasets if available, because of the limitation of the prompt size. Furthermore, we assume the availability of high-quality LLMs, which might not be the case for some low-resource languages.

\section{Ethical Statement}
LLMs are prone to hallucination. In our case, the extracted evidence could be incorrect due to hallucination. Furthermore, the prompts can be tweaked to intentionally generate wrong evidence or predictions. We caution the reader to be aware of such issues and to not misuse the system.

\bibliography{custom}

\end{document}